\documentclass[sigconf]{acmart}
\AtBeginDocument{%
  }

\acmConference[ICONS '23]{On-Sensor Data Filtering using Neuromorphic Computing for High Energy Physics Experiments}{August 01--03, 2023}{Santa Fe, NM}
\usepackage{soul}

\usepackage{algorithm}
\usepackage{algorithmic}
\usepackage{graphicx}
\usepackage{subfig}
\begin{document}

\title{On-Sensor Data Filtering using Neuromorphic Computing for High Energy Physics Experiments}

\author{Shruti R. Kulkarni, \\Aaron Young, Prasanna Date, \\Narasinga Rao Miniskar, \\ Jeffrey S. Vetter}
\affiliation{%
  \institution{Oak Ridge National Laboratory}
  \streetaddress{P.O. Box 1212}
  \city{Oak Ridge}
  \state{Tennessee}
  \country{USA}
  \postcode{}
}

\author{Farah Fahim, \\Benjamin Parpillon, \\ Jennet Dickinson, Nhan Tran}
\affiliation{%
  \institution{Fermi National Accelerator Laboratory}
 % \streetaddress{1 Th{\o}rv{\"a}ld Circle}
  \city{}
  \country{USA}}
\email{}

\author{Jieun Yoo,\\  Corrinne Mills}
\affiliation{
\institution{University of Illinois}
\city{Chicago}
\country{USA}
}

\author{Morris Swartz, Petar Maksimovic}
\affiliation{
\institution{John Hopkins University}
\city{}
\country{USA}
}

\author{Catherine D. Schuman}
\affiliation{%
  \institution{University of Tennessee}
  \city{Knoxville}
  \country{USA}
}

\author{Alice Bean}
\affiliation{%
  \institution{University of Kansas}
  \city{}
  \country{USA}}
\email{}

\renewcommand{\shortauthors}{Kulkarni et al.}

\begin{abstract}
In this paper, we investigate the prospects of  applying neuromorphic computing spiking neural network models to filter data on the readout electronics of the sensor in the high energy physics experiments at the High Luminosity Large Hadron Collider. We present our approach  on developing a compact neuromorphic model that filters out the sensor data based on the particle's transverse momentum with the goal of reducing the amount of data being sent to the downstream electronics. The incoming charge waveforms are converted to streams of binary-valued events which are processed by the SNN. %We present our insights on the various model development choices from data encoding to  optimization of the training algorithm for the purpose of developing a fast and compact model that enables accurate processing the incoming temporal data and is also optimized for hardware deployment. 
We present our insights on the various system design choices from data encoding to optimal hyperparameters of the training algorithm for an accurate and compact SNN optimized for hardware deployment. Our results show that an SNN trained using an evolutionary algorithm and with optimized set of hyperparameters shows a signal efficiency of about $91$\% with nearly half the number of parameters than a Deep Neural Network.
\end{abstract}




\received{30 April 2023}
\received[revised]{}
\received[accepted]{}

\maketitle

\section{Introduction}

Neuromorphic computing has emerged as a promising computing paradigm for several edge and smart-sensor processing applications where energy efficiency and area constraints are the key requirements \cite{schuman2022opportunities,patton2022neuromorphic,aimone2022review}. Neuromorphic computing uses networks of bioplausible neurons, or spiking neurons that compute with binary valued signals (called spikes). The networks of these neurons, called Spiking Neural Networks (SNNs) have an inherent notion of time embedded in their dynamics as synaptic delays and neuronal time constants \cite{date2022neuromorphic,date2021computational}. SNNs have been demonstrated for very diverse sets of cognitive and non-cognitive applications such as autonomous navigation, anomaly detection, graph algorithms, epidemic modeling, etc. \cite{patton2021neuromorphic,date2018efficient,cong2022semi,hamilton2020modeling,schuman2022evoimitation}.

To develop a complete neuromorphic system, in addition to implementing the SNN training algorithm, a key challenge to be addressed is the best spatial and temporal data encoding schemes needed to encode the input data before feeding it to the SNNs. There have been several spike encoding schemes proposed in the literature for converting real-valued signals into streams of spikes \cite{schuman2019non, wang2022efficient}, and those that are based on the design specifications of the sensory hardware such as dynamic vision sensors, and tactile sensors that generate a stream of events in response to the sensed signal \cite{Gallego2022event, birkoben2020spiking}.  In this paper, we study the prospective of deploying a neuromorphic solution for data filtering in the design of the pixel detectors for high energy particle physics experiments, where the area and power requirements of the hardware detector are heavily constrained.

High energy physics experiments at the Large Hadron Collider (LHC)
rely on complex detectors with over a billion detector channels to make precise measurements of proton-proton collisions occurring at a frequency of 40MHz. These detectors generate data at rates on the order of few Peta bytes per second.  In order to reduce the collected data to a manageable size, the current generation of LHC experiments apply physics-motivated selection by means of a two-tiered trigger system.  The first trigger level (level one) is composed of custom hardware processors that save collision data at a rate of around 100 kHz. The second level, known as the high-level trigger, consists of a farm of processors that reduces the event rate to around 1kHz before data storage.

Due to the large number of read out channels in a pixel detector, information from these sub-systems has thus far only been employed at the second level of the trigger system. However, machine-learning based selection implemented directly in the pixel detector front-end electronics has the potential to reduce the size of pixel detector data, and could provide a path to using the physics information collected by this sub-system in the level one trigger selection. Implementing such an algorithm in the detector front-end electronics requires that the processing models are compact enough to be deployed in a small area and with low power, and can operate with a latency below $200\,$ns. There have been several demonstrations of custom ASICs and FPGA designs as accelerators for implementing machine learning models for sensor data filtering, where latency is a key criteria along with the model's algorithmic performance 
\cite{khoda2022ultra,di2021reconfigurable}. 
Recent works have also illustrated that analog in-memory computing systems can support the high bandwidth requirements as in HEP applications, where neuromorphic computing has great potential in meeting the system constraints \cite{kosters2023benchmarking}.
% \cite{kosters2023benchmarking} have shown analog-in-memory can potentially achieve high bandwidth for HEP and other scientific applications, also made a case for neuromorphic computing...

% The SNN approach has great potential to perform efficently within these constraints.

 In this work, we make use of the evolutionary computing algorithm -- Evolutionary Optimization for Neuromorphic Systems (EONS) \cite{schuman2020evolutionary}, to train SNNs which has been used for prior work on the application of SNNs for scientific applications \cite{schuman2017neuromorphic}. It was shown there that EONS was able to generate very compact SNNs with the same accuracy as DNNs. Hence, here we use it as the training algorithm for the smart-pixels application. The overall processing pipeline from data encoding to training has several hyperparameters that need to be tuned. For this optimization process, we make use of a design space exploration tool called DEFFE (data efficient exploration framework) \cite{liu2020deffe}, as discussed in the following sections. %We also use the DEFFE framework to perform a search of the hyperparameters to obtain the optimal configuration for the end-to-end system.

Section 2 presents an overview of the pixel detection system, and the process of on-chip detection of the particle hits and inferring their physics information from the sensor values. Section 3 goes over the neuromorphic process of filtering the sensor array data based on the particle momentum values. Section~\ref{sec:results} presents the results of our studies. Finally, in Sections \ref{sec:disc} and \ref{sec:summary} we present our insights from our studies on opportunities and challenges for deploying neuromorphic systems for such area and energy constrained applications.

\section{Smart-pixel System}
\label{sec:smart-pixel}
In high energy collider experiments, the sub-system of the detector nearest to the collision point is typically a tracking detector composed of concentric layers of pixelated silicon sensors.  When a charged particle traverses a pixel detector, the pattern of charge deposited in each silicon layer can be used to reconstruct the particle's trajectory, or track. These particle tracks provide high precision measurements of the direction and origin (vertex) of each particle. In addition, the pixel detector is immersed in a strong magnetic field parallel to the beam, which causes particle tracks to curve. The radius of curvature of a charged particle track is directly proportional to the particle's momentum in the plane transverse to the beam ($p_T$).  

Charged particles with high $p_T$ form a component of most collision events containing heavy or high energy particles. Such signatures are of primary interest in collider experiments such as ATLAS and CMS \cite{atlas,cms}. Events containing only low $p_T$ particles, on the other hand, typically result from low energy physics processes and are generally discarded by the trigger.  

\begin{figure}[htbp]
  \centering
  \includegraphics[width=0.3\textwidth]{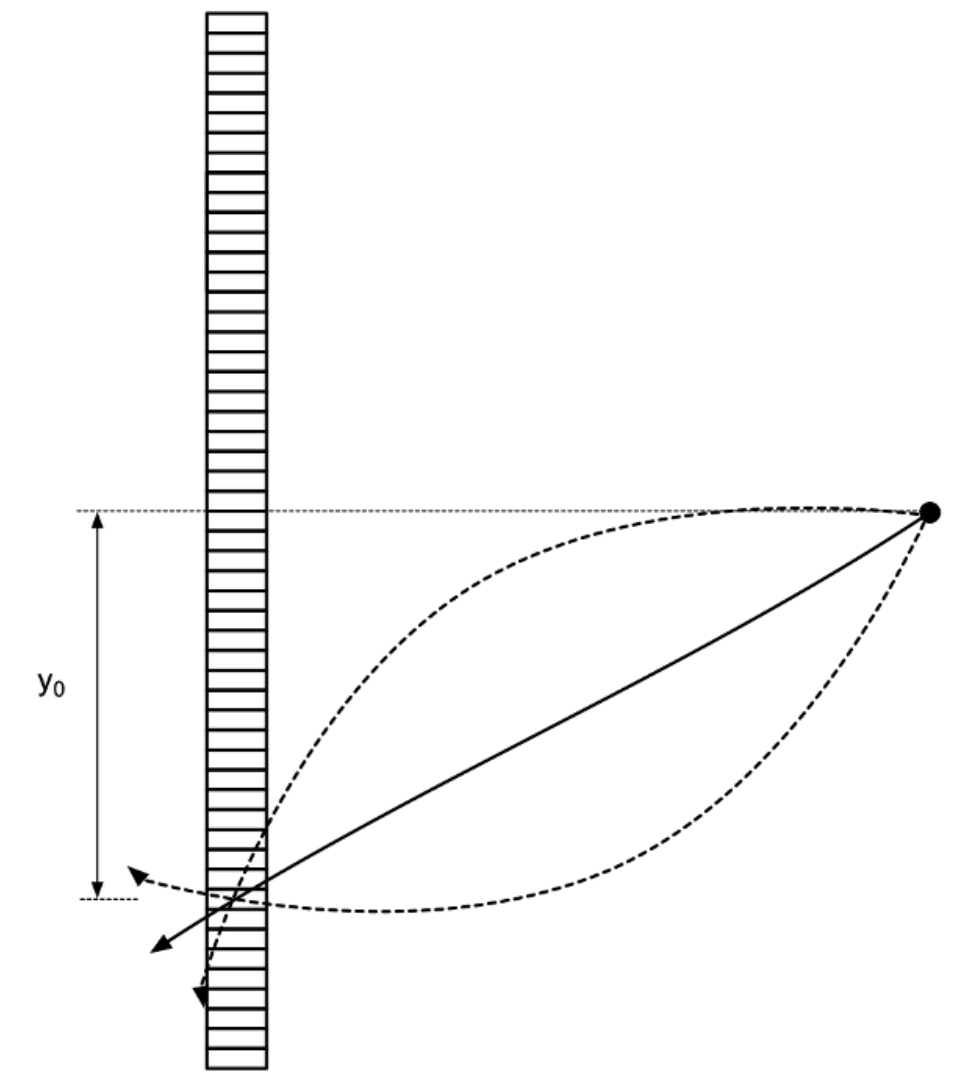}
   \caption{Diagram of three charged particle tracks traversing a sensor. The sensor is viewed in the bending plane of the magnetic field. The solid track corresponds to a charged particle track with high $p_T$ , while the two dashed tracks correspond to low $p_T$ tracks from particles with opposite charge.}
  \label{fig:pt-cartoon}
\end{figure}

The pattern of charge deposited in a pixel sensor by a traversing particle is highly correlated with the curvature of the track and therefore with its $p_T$. This is demonstrated in Figure \ref{fig:pt-cartoon}, which shows an edge-on view of a silicon sensor and three charged particle tracks. The straight track corresponds to a particle with high $p_T$, while the two curved dashed tracks correspond to low $p_T$ tracks from particles with opposite charge. 

Charged particles with $p_T<2$ GeV make up more than 90\% of all tracks recorded in proton-proton collisions at center of mass energy 13 TeV. Filtering out low $p_T$ tracks on-detector using information from a single pixel layer has the potential to dramatically reduce the amount of data read out by the pixel detector, provided it maintains high efficiency on high $p_T$ tracks and fits within detector constraints. A compact machine learning classifier is therefore trained to separate charge clusters based on particle $p_T$.

% \textbf{Details about related works:} Machine Learning -- Deep Neural Network models being used for sensor data filtering with custom ASICs or FPGA AI accelerators... latency is a key criteria along with the model's algorithmic performance.
% \cite{khoda2022ultra,di2021reconfigurable}

% \cite{kosters2023benchmarking} have shown analog-in-memory can potentially achieve high bandwidth for HEP and other scientific applications, also made a case for neuromorphic computing...

%  \begin{itemize}
%     \item Role of AI  - DNN and Neuromorphic models
% \end{itemize}

\subsection{Datasets for Training}

\begin{figure}
    \centering
    \includegraphics[width=0.4\textwidth]{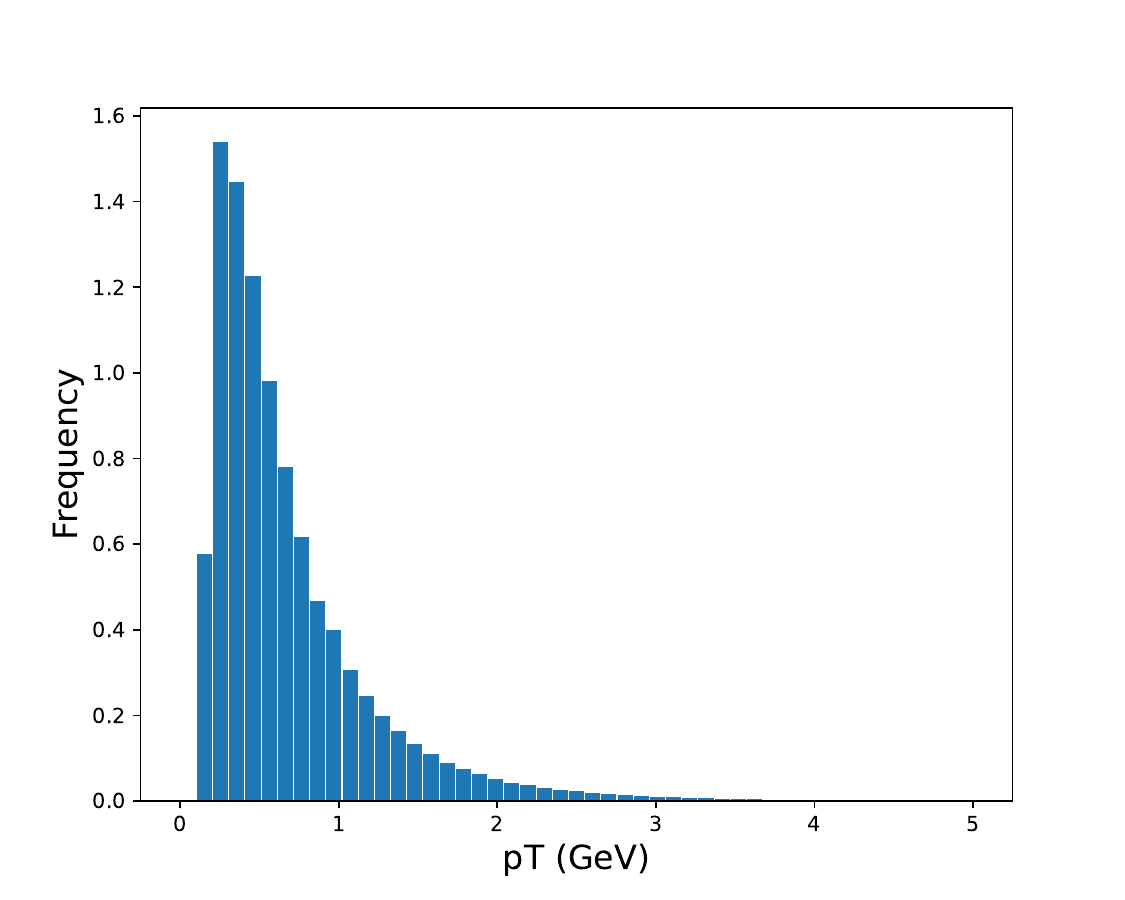}
    \caption{Distribution of pT values of the clusters within the dataset of about four million samples.}
    \label{fig:pt}
\end{figure}

A simulated dataset of four million charged particle interactions in a silicon pixel sensor has been produced with particle kinematics taken from tracks measured by the CMS experiment ~\cite{zenodo}. Figure \ref{fig:pt} shows the $p_T$ spectrum of the incident particles (pions). 
Because few particles with very low $p_T$ are reconstructed as tracks in CMS, the track $p_T$ distribution begins around 200 MeV.

The simulated sensor measures 1.6x1.6 cm${^2}$ with a thickness of 100 $\mu$m. The sensor plane is described by coordinates $x$ and $y$, and the pixel pitch is 50 taken to be 50x12.5$\mu$m in $(x,y)$.  Each cluster in the sensor is contained in a region of interest that spans 21x13 pixels centered at position $(x_0, y_0)$.  The sensor is positioned 30mm from the particle's origin point. The detector is immersed in a 3.8\,T magnetic field parallel to the $x$ coordinate. The pattern of charge deposited by each particle in the pixel sensor is simulated in 200 picosecond steps using a time-sliced version of PixelAV~\cite{pixelav}.

For this work, we have restricted our training samples to that of the particle tracks traversing the central 2 mm of the sensor in the $y$ direction (-1 $<y_0<$ 1 mm).  We also consider only positively charged particles in the training. % this is what is meant by the sign on pT

% \begin{itemize}
%     \item Over view of the data encoding scheme
%     \item Description of the parameters at each stage of the processing pipeline
% \end{itemize}

\section{Method}
\label{sec:nmc}
\begin{figure}
    \centering
    \includegraphics[width=0.47\textwidth]{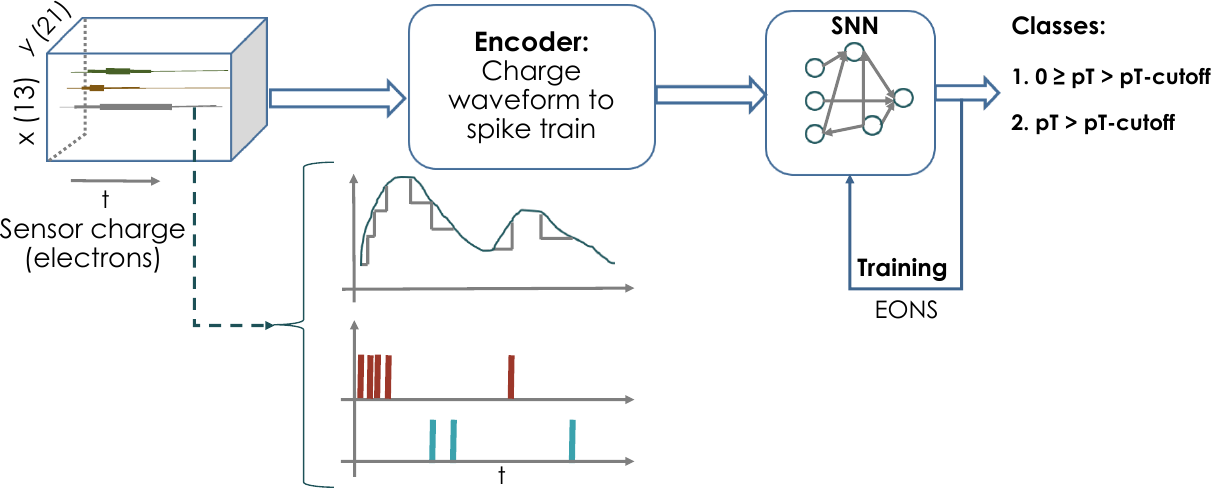}
    \caption{End-to-end representation of the sensor data encoding and processing with SNNs. Each of the sensor pixel in the $13 \times 21$ array is encoded into a stream of spikes into two channels, one for capturing the timing in the rising edge and second for the timing capture in the falling edge of the incoming signal.}
    \label{fig:smart-pixels}
\end{figure}

We employ the principles of neuromorphic computing to design an efficient data filtering model for the smart-pixels system for the HEP experiments. Figure~\ref{fig:smart-pixels} shows the end-to-end processing pipeline for classifying the sensor data into high $p_T$ or low $p_T$ clusters with neuromorphic computing. As seen in the figure, the sensor charge waveform first needs to be first converted into spike trains and applied to SNN. As an SNN has multiple parameters such as synaptic weights and delays and neuronal thresholds, we employ the evolutionary algorithm called Evolutionary Optimization for Neuromorphic Systems (EONS) to train the network. The key criteria in designing a neuromorphic model is the computation of the dynamics of the bio-plausible models of neurons and synapses, and the communication mechanism implemented using binary valued events called spikes.  The following sub-sections present the details of our simulation framework, encoding and decoding mechanisms and the training approach.

\subsection{Simulation Environment}
We use the network of simple leaky-integrate-and-fire neurons and synapses with weights and delays to realize the classification model for the smart-pixel detector. The target neuromorphic hardware used to realize the SNNs is the FPGA based Caspian device \cite{mitchell2020caspian}, which uses integer representation for all of the network parameters. Caspian is an event based low-power neuromorphic hardware suitable for edge applications.

For our current work, we use the software simulator of Caspian within the neuromorphic TENNLab software framework \cite{plank2018tennlab}. The framework allows the users to define the graphs of the SNNs, the input spike events to be applied to the network and finally load them on to the required simulator, which in the current work is the Caspian simulator. The framework also provides various output spikes decoding strategies so that the network output can be interpreted according to the application \cite{schuman2019non}. We set the precision of synaptic weights to signed integer of $9-$bits, delays to $4-$bits, and neuronal thresholds to $8-$bits. We explore the different input spike encoding and data reduction schemes as discussed in the following sub-sections. The output of the network is decoded based on the output neuron that fires the highest as defined within the TENNLab framework.

\label{sec:overview}

\subsection{Data Encoding}
\begin{figure}
    \centering
    \includegraphics[width=0.4\textwidth]{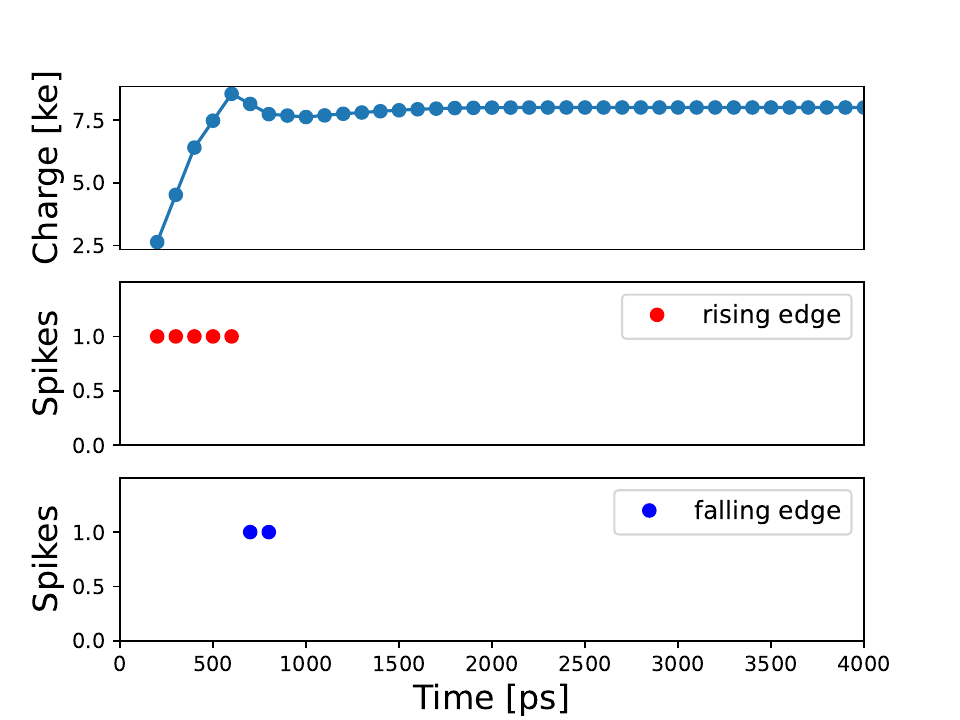}
    \caption{Each of the charge waveform (amplitude measure in electrons) from the $13\times 21$ pixel array is converted into a stream of spikes with inter-spike intervals decided by the rising or falling time of the waveform. The spikes shown in red represent the rising edge of the waveform, while those in blue represent the falling edge. The charge data is up-sampled to have a temporal resolution of $100\,$ps.}
    \label{fig:sensor-signal-spikes}
\end{figure}

The sensor data in the dataset consists of frames of charge values (in electrons) for 4000 ps, sampled at every 200\,ps. For the neuromorphic classifier, we convert the charge waveforms into trains of spikes, with the spike times encoding the rise times or fall times of the incoming waveform. To study the impact of time-resolution in the data encoding phase, we also incorporate linear up-sampling of the data to resolution below $200\,$ps. Each input pixel in the $13\times 21$ array of the sensor data is represented by two channels of spike trains, one for the rising edge of the waveform and other for the falling edge. Figure~\ref{fig:sensor-signal-spikes} shows how an input charge waveform (top panel) is converted into rising edge spikes (middle panel) and falling edge spikes (bottom panel). Similar neuromorphic data encoding mechanisms have been employed in various event based sensors such as the Dynamic Vision Sensor cameras, tactile sensors, etc. \cite{Gallego2022event, zeng2021neuromorphic}.

% We implement a software module to perform the spike encoding with the specifications that would be used  on the front-end hardware specifications. We apply a threshold of 800e$^-$ on the sensor charge data to mimic the noise suppression. The analog data is then converted into spike trains based on the successive charge differences in the waveform (see Figure~\ref{fig:smart-pixels}). We use a difference of $400\,e^-$ to generate a spike in the positive or negative channel. 

We implement a software module to perform the spike encoding based on the temporal aspects of the incoming sensor charge. We first apply a threshold of 800\,e$^-$ on each of the pixels in the cluster frame to mimic the noise suppression. We then compute the successive differences in the temporal data, and generate a spike whenever there is a difference of $400\,e^-$ in the incoming time series (see Figure~\ref{fig:smart-pixels}). Hence, a fast rising (or falling) waveform would create more closely spaced spikes, while a slow rising (or falling) waveform would generate spikes farther apart in time. Algorithm~\ref{algo:charge2spike} presents our encoding approach for converting time series of charge values into a spike train. Here $\mathbf{x}$ is the timeseries input of charge values per pixel, $x_{th}$ is the baseline threshold to mimic the noise suppression, and $\Delta x$ is the difference in the charge values to generate a spike.

\begin{algorithm}
\caption{Spike Encoding of Charge Values}
\label{algo:charge2spike}
\begin{algorithmic} 
\STATE  $\mathrm{Encoder} (\textbf{x}, x_{th}, \Delta x)$
% % \INPUT $x_0, x_1, ..., x_N$
% % \INPUT $x_{th}$
% % \INPUT $\Delta x$
 \STATE $k=0$
 \STATE $t_{+} = []$, $t_{-}=[]$
\WHILE{ $k < N$} 
% \STATE $y \leftarrow 1$
\IF{$x_k > x_{th}$}
\STATE $x_{rise} = x[k] + \Delta x$
\STATE $x_{fall} = x[k] - \Delta x$
\STATE $p = argmin([x_{rise}-x[k+1:]])$
\STATE $q = argmin([x_{fall} - x[k+1:]])$
\IF{$p>q$}
\STATE $k = k+q$
\STATE $t_{-}.append(t_{res} \times (k+1))$
\ELSE
\STATE $k = k+p$
\STATE $t_{+}.append(t_{res} \times (k+1))$
\ENDIF
\ENDIF
 \ENDWHILE
 \STATE return $t_{+}$, $t_{-}$
% \ENDWHILE
\end{algorithmic}
\end{algorithm}

% \begin{itemize}
%     \item Hardware for the dvs like events generation
%     \item Simulation of the hardware -- converting data into spikes based on charge difference
%     \item Up- scaling the charge time-slices to avoid data loss
%     \item hyperparameter of encoding process - timescale of data sampling
%     \item Two channels per input pixel of the sensor array - rising and falling edge pulses/spikes
%     \item Figures showing examples
% \end{itemize}

\subsection{Spatial Data Reduction}
There are often advantages gained from reducing the input space when working with SNNs. 
Specifically, when training SNNs with evolutionary optimization some of the benefits can include smaller network sizes, reduced per epoch training time, and faster increase in training performance.
These benefits come from a smaller number of input neurons and the reduced complexity generating smaller networks leveraging this input. 

When designing input reduction methods, it is important to try to avoid losing important features in the input as the input space is reduced.
Additionally, reductions that could help the algorithm converge to a more generic solution are advantageous.
With this in mind, we developed multiple spatial data reduction patterns to try as part of our hyperparameter search.
From looking at the training data, active pixels are clustered around the center of the image, and the cluster is a local region of activity with the majority of with no activity.
The particle hits that generate the clusters of charged pixels are sparse enough that only one cluster is likely to be present in the image at a time.
The centered cluster activity is an artifact of how the training data was generated, and in a hardware deployment, the clusters would not be centered.
Therefore, we thought that a spatial reduction pattern, which preserved local cluster details and lost x, y pixel coordinate details, would still retain the crucial information needed to classify the $p_T$ while making the solution more generic.

\begin{figure}
    \centering
    \includegraphics[width=0.45\textwidth]{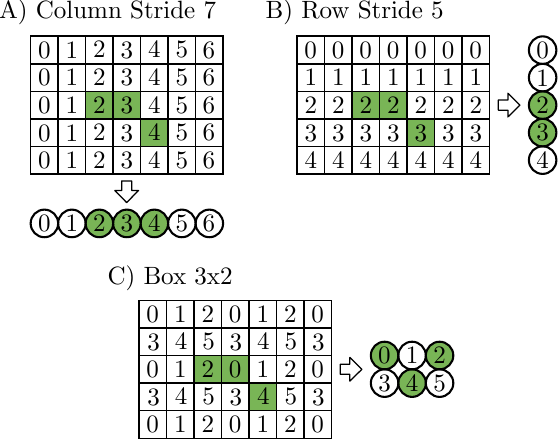}
    \caption{Examples of different spacial data reduction patterns applied on a 7x5 image. A) shows the column stride 7 pattern for a column-wise reduction, B) shows the row stride 5 pattern for a row-wise reduction, and C) shows the box $3\times2$ pattern. The boxes represent image pixels and the circles are input neurons. The green highlights represent spikes in the input.}
    \label{fig:pattern_reduction}
\end{figure}

For all of the special reduction patterns, a repeating numerical pattern is applied to each pixel in the input image, and each pixel with the same number is reduced to a single input neuron using an ``OR'' operation.
The pattern generators we implemented are column stride, row stride, and box.
The row and column stride patterns take a stride length and count to that number along the rows or columns, respectively.
When the stride length matches the dimension of the image, a column-wise or row-wise reduction is performed.
The box pattern tasks the size of the box along the x and y dimensions.
The numbers then count along in that square before repeating in the neighboring squares.
The size of the box can be adjusted to capture larger local regions.
Examples of these reduction patterns are shown in Figure~\ref{fig:pattern_reduction}.
For the hyperparameter search, we explored row-stride 13 (row-wise reduction), row-stride 26, stride 21 (column-wise reduction), stride 42, and multiple box sizes ranging in edge lengths between 2 and 8.
These patterns were chosen to explore how much local information is needed to correctly classify high or low $p_T$.

% Comparison with y-profile feature used in DNN

\subsection{EONS}

The training approach that we leverage in this work is Evolutionary Optimization for Neuromorphic Systems (EONS)~\cite{schuman2020evolutionary}.  EONS optimizes the structure and parameters of spiking neural networks for neuromorphic deployment.  It optimizes SNNs within the constraints of the underlying hardware platform, including parameter precision constraints. EONS is an evolutionary approach that begins with a collection of randomly initialized potential SNN solutions to form its initial population.  The initial population may also be seeded with SNNs evolved in previous EONS runs, created through another training approach, or hand-tuned by a user. Once an initial population is formed, that population is evaluated and each network is given a fitness score.  The fitness scores are used to drive tournament selection and reproduction operations that produce a child population.  The reproduction operations include cloning, crossover, and random mutation. EONS also uses an elitism mechanism to maintain the best networks from the previous population in the child population.  This evaluation, selection, and reproduction process is repeated until a stopping criteria (e.g., maximum number of generations or desired fitness score) is reached.

\subsection{SNN Training}
We carry out all the SNN training and optimization within the TENNLab framework. The dataset consists of over four million samples, with samples divided into $160$ files, with each file having on an average $25,000$ samples. SNNs are trained with EONS to classify samples into high $p_T$ and low $p_T$. The inputs to the SNN are fixed by the number of channels of spikes coming in from the encoder. The number of output neurons of the network is always set to two, one corresponding to the low $p_T$ bin and other for high $p_T$. The division of the samples into high and low $p_T$ clusters is done using a $p_T$ cut-off value, which is one of the hyperparameters. The training-test split is done on the dataset files, and each epoch of training uses the data from a randomly chosen training set file. As the dataset is not uniformly distributed across the entire range of $p_T$ (see Figure~\ref{fig:pt}), during training we balance the number of samples in each of the two bins, i.e., high $p_T$ and low $p_T$ classes. 

We also explore the different fitness functions to guide the training process with EONS -- classification accuracy and penalty. Classification accuracy measures the fraction of samples correctly classified into their correct classes. We use the accuracy score provided by the scikit learn python library.
The penalty-based score is computed by giving a negative score to the samples that are incorrectly classified. The score is also weighted based on how far is the actual $p_T$ from the defined classifier $p_T$ cut-off point using a $\tanh$ function. The penalty on the training set is computed as:
\begin{equation}
    s = - \sum_i E_i \times (\tanh(k\times |p_{Ti} - p_{Tth}|))
\end{equation}

Here, $E = y_o - y_t$ is the difference between the predicted ($y_o$) and actual ($y_t$) bin of an input sample. The parameter $k$ is used to control the slope of the tanh curve. The score is set as negative, since EONS tries to maximize the fitness value.

During test, we evaluate the network's classification performance on the entire data from the test-set files. In addition to the model's classification accuracy and f1 score, we also evaluate the model's signal efficiency and data reduction abilities as discussed in Section~\ref{sec:smart-pixel}. Signal efficiency is calculated as the number of samples correctly identified as high $p_T$, whose true $p_T$ lies above $2\,$GeV. Data reduction is calculated as the number of samples correctly identified as low $p_T$, whose true $p_T$ lies below $2\,$GeV.
% Why do we use a different pT cut-off in training vs test?

% \begin{itemize}
%     \item Details about EONS \cite{schuman2020evolutionary}
%     \item Challenge with the dataset size -- partial data loading each epoch
%     \item Fitness fucntions for EONS - accuracy, penalty
% \end{itemize}

\subsection{Hyperparameter Optimization}
The hyperparameter optimization of the training process is achieved with the DEFFE framework. DEFFE \cite{liu2020deffe} is a data-efficient exploration framework originally designed for design space exploration of architecture parameters with evaluated cost metrics for each parameter set sampled. It is modular with pipelined stages composed of a sampling stage to create sets of samples out of large search space, an evaluate stage to evaluate the set of samples in using parallel processing on a single system or using a SLURM environment on a compute cluster, and lastly an extract stage to extract the desired cost objects out of the evaluated samples. It has optional pipeline stages, such as a machine learning stage which trains the configured machine learning model (random forest or convolution neural network) with already evaluated samples to predict the cost objects, which can be used in the sampling stage to decide the next set of samples for evaluation. DEFFE supports a wide range of sampling techniques such as random sampling, machine learning-based sampling, and DOEPY (Design of Experiment Generator in Python) sampling techniques such as lattice, hypercube, etc.
On the infrastructure side, DEFFE provides a scalable computing platform to further reduce the runtime needed for performance estimation by harnessing the parallelism of HPC (High-Performance Computing) clusters. The net outcome is the runtime of a typical performance modeling task, which could take a few months on a single node computer, is reduced to a few days by using the ML-learning method and further reduced to a few hours when the simulation is executed on a 20-node cluster. 

There are many design parameters that are driven by different levels of the Spiking Neural Network design stack. DEFFE automates the process of analyzing these design parameters, extending our ability to explore the trade-offs of various design options. For example, in the case of exploring SNN Classification discussed prior, DEFFE is configured with knobs for the conversion of the simulated dataset's charge values to spike pulses, knobs for the hyperparameters for the EONS training process, and knobs for the design and configuration of the neuromorphic processor. Then the DEFFE infrastructure can evaluate and collect relevant metrics for each of these configurations by calling our training tools with each configuration and then parsing the output to collect metrics from that configuration. DEFFE evaluates the specific configurations in parallel and collects the results into a single table. By using DEFFE, the setup of experiments is greatly simplified. The DEFFE configuration file specifies the knobs which can be adjusted and the metrics to be collected. Scripts are added and used by DEFFE to take a specific configuration and evaluate that configuration. Then, DEFFE handles the launching of each job on a high-performance-compute cluster to explore the design space. 

\section{Results}
\label{sec:results}
% Details about parameters kept fixed based on initial runs -- no. of epochs, EONS parameters, data encoding parameters (timescale, charge difference, etc.

% and parameters explored by Deffe
We carried out the EONS training runs using DEFFE, which launched the evaluation jobs in parallel on four Linux servers, each with two AMD EPYC 7742 64-Core processors and 1TB of RAM.
Table~\ref{tab:hyp} lists the  hyperparameters for our neuromorphic training that are optimized with DEFFE. For EONS, we keep the SNN population size as $100$, starting nodes at $50$, and starting edges at $800$ based on our initial experiments. We also parameterize the time resolution of encoding the charge waveform to spike pulses (see Figure~\ref{fig:sensor-signal-spikes}), which also decides the number of timesteps needed for the SNN evaluation. Our search space also includes a bias input, which provides spike pulses at a constant rate, which EONS uses to make useful connections within the SNN if it helps improve the fitness score. 

\begin{table}[]
\caption{Hyperparameters for SNN Training}
  \label{tab:hyp}
\begin{tabular}{ll}
\hline
\textbf{Hyperparameters}        & \multicolumn{1}{c}{\textbf{Values}}                                                                                                                                                                          \\ \hline
\textbf{Timescale}              & [10, 20, 40, 50, 200]                                                     \\
\textbf{Spatial data reduction} & \begin{tabular}[c]{@{}l@{}}[full image (13 x 21), row-stride 13, \\ row-stride 26, stride 21, stride 42, \\ box 2 2, box 3 3, box 4 4, box 2 4, \\ box 4 2, box 2 8, box 8 2 box 4 8, \\ box 8 4]\end{tabular} \\
\textbf{pT cut-off}             & [0.2, 0.5, 0.7]                                                                  \\
\textbf{EONS fitness}           & [Accuracy, penalty, combination]                                                                                                         \\
\textbf{Bias}                   & [True, False]                                                       \\ \hline
\end{tabular}
\end{table}

\subsection{SNN results}
\begin{figure}[!htb]
\subfloat[]{
\includegraphics[width=0.48\linewidth]{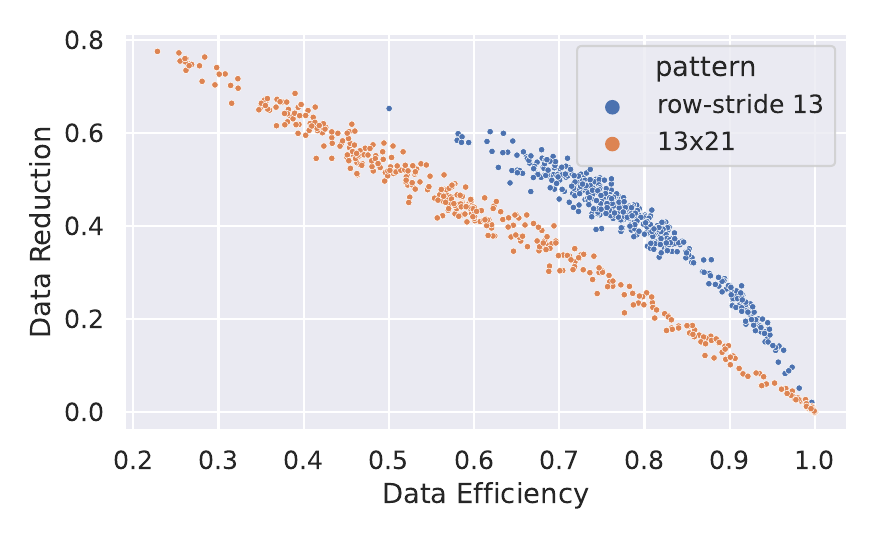}
} \hfill
\subfloat[]{
\includegraphics[width=0.48\linewidth]{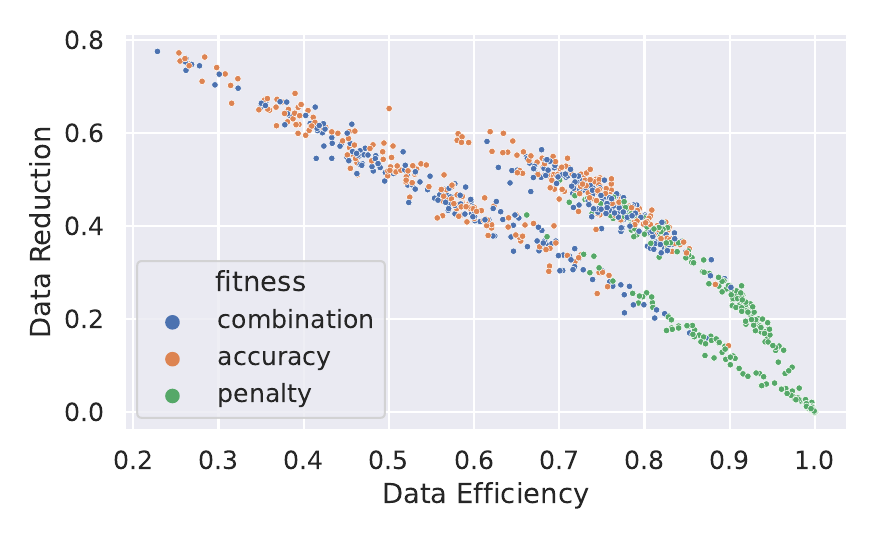}
}\\
\subfloat[]{
\includegraphics[width=0.48\linewidth]{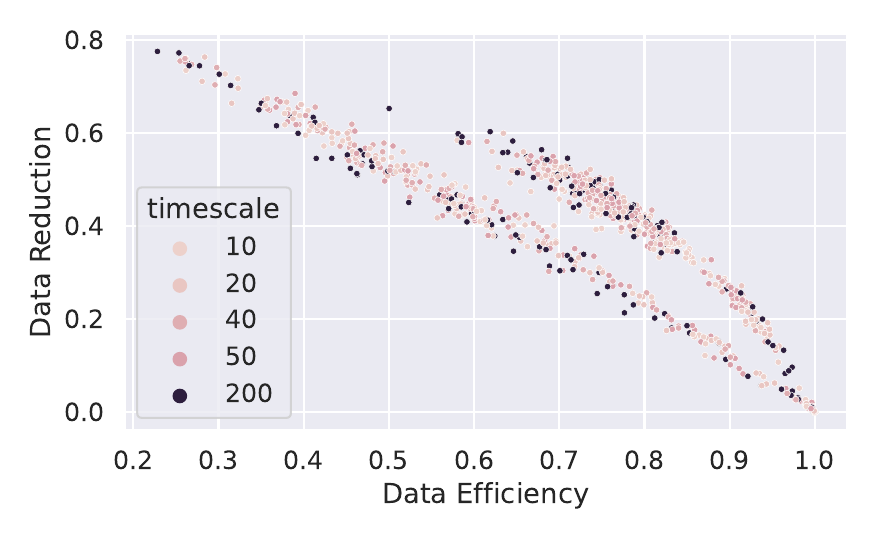}
} \hfill
\subfloat[]{
\includegraphics[width=0.48\linewidth]{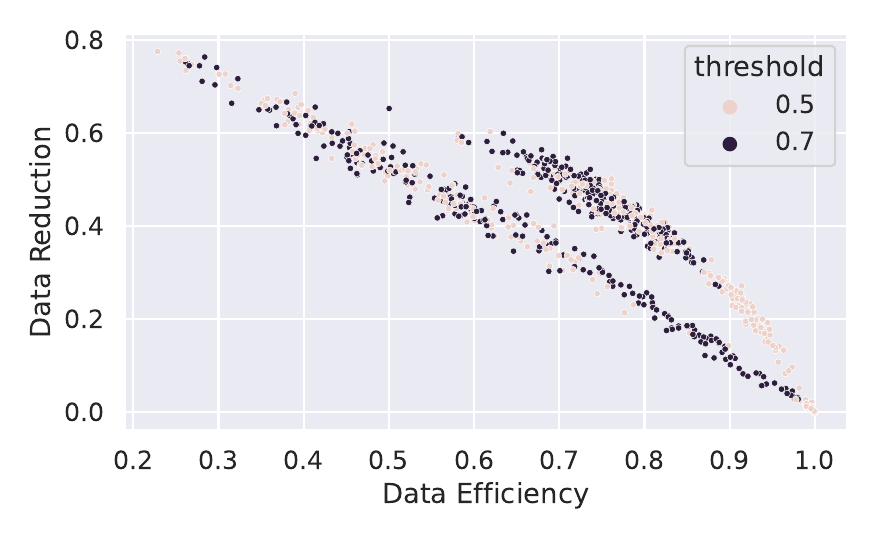}
}
 \caption{Signal efficiency and data reduction as a function of different encoding and training hyperparameters -- (a). Spatial data reduction strategies, (b). Fitness functions during EONS training, (c). Time resolution used in spike encoding, and (d). pT-cutoff during training.}
    \label{fig:fitness}
\end{figure}

We started by running DEFFE with $120$ different hyperparameter combinations from the ones listed in Table~\ref{tab:hyp}. These runs were carried out with $500$ epochs of EONS training. Figure~\ref{fig:fitness} shows the trend between signal efficiency and data reduction metrics of different runs impacted by different input and EONS hyperparameters. %It can be seen that with signal efficiency of $91.0\%$, the best data reduction is $26.0\%$. From the top 
As seen in Figure~\ref{fig:fitness}(a), it can be seen that reducing the data spatially significantly improved the performance compared to the full two-dimensional data frame of $13\times21$ sized array. 
We also note that the time-resolution did not make any notable difference in the performance. %Hence, we carried out our further experiments with time resolution of $200\,$ps, which results in $20$ timesteps for the SNN's evaluation. 
We evaluated three different fitness functions for EONS -- accuracy, penalty and a combination of the two. The combination fitness evaluated with penalty score for the first half of training and then used accuracy for the remaining half. It can be seen that penalty based scoring was favored by higher efficiency networks.  The $p_T$ cut-off value also needs to be at an optimal value for better performance of the SNN. The goal of the classifier model is to reject as many low $p_T$ samples as possible. With $p_T$ cutoff of $0.2$, we observed that the SNN was not seeing enough of low-$p_T$ clusters during training, which higher $p_T$ lead to decrease of signal efficiency. From the figure we see that at $0.5$ the model was able to achieve higher efficiency.

% \begin{figure}[!htb]
% \begin{minipage}[t]{0.24\textwidth}
% \centering
% \includegraphics[width=0.93\textwidth]{Neuro-synaptic_core_blocks.pdf}
% \end{minipage}
% \hfill
% \begin{minipage}[t]{0.24\textwidth}
% \centering
% \includegraphics[width=1.0\textwidth]{sttram_core.pdf}
% \end{minipage}

% \begin{figure*}[t!]    
% \subfloat[A Cap]{%
%             \includegraphics[width=.48\linewidth]{example-image-a}%
%             \label{subfig:a}%
%         }\hfill

\begin{figure}
    \centering
    \includegraphics[width=0.5\textwidth]{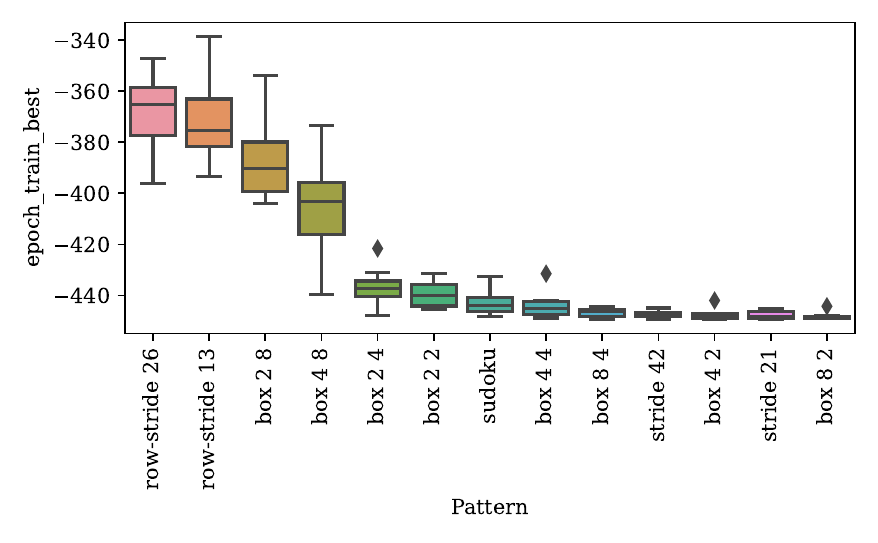}
    \caption{EONS penalty-based fitness score as a function of different spatial data reduction schemes.}
    \label{fig:spatial_enc}
\end{figure}

As discussed above, reducing the data spatially showed an improvement over the full two-dimensional frame, we further trained with different spatial data reduction schemes. Figure~\ref{fig:spatial_enc} shows the training fitness scores (penalty-based) across the different encoding schemes. It can be seen that the y-dimension of the input array has a strong impact on the performance of the SNN classifier. %The `row-stride' encoding schemes project the input 2D array along the columns, while the `box 2 8' and `box 4 8' have project the 2D array of spikes into rectangular windows where the y-dimensional is larger than the x. 

\begin{figure}[b]
    \centering
    \includegraphics[width=0.45\textwidth]{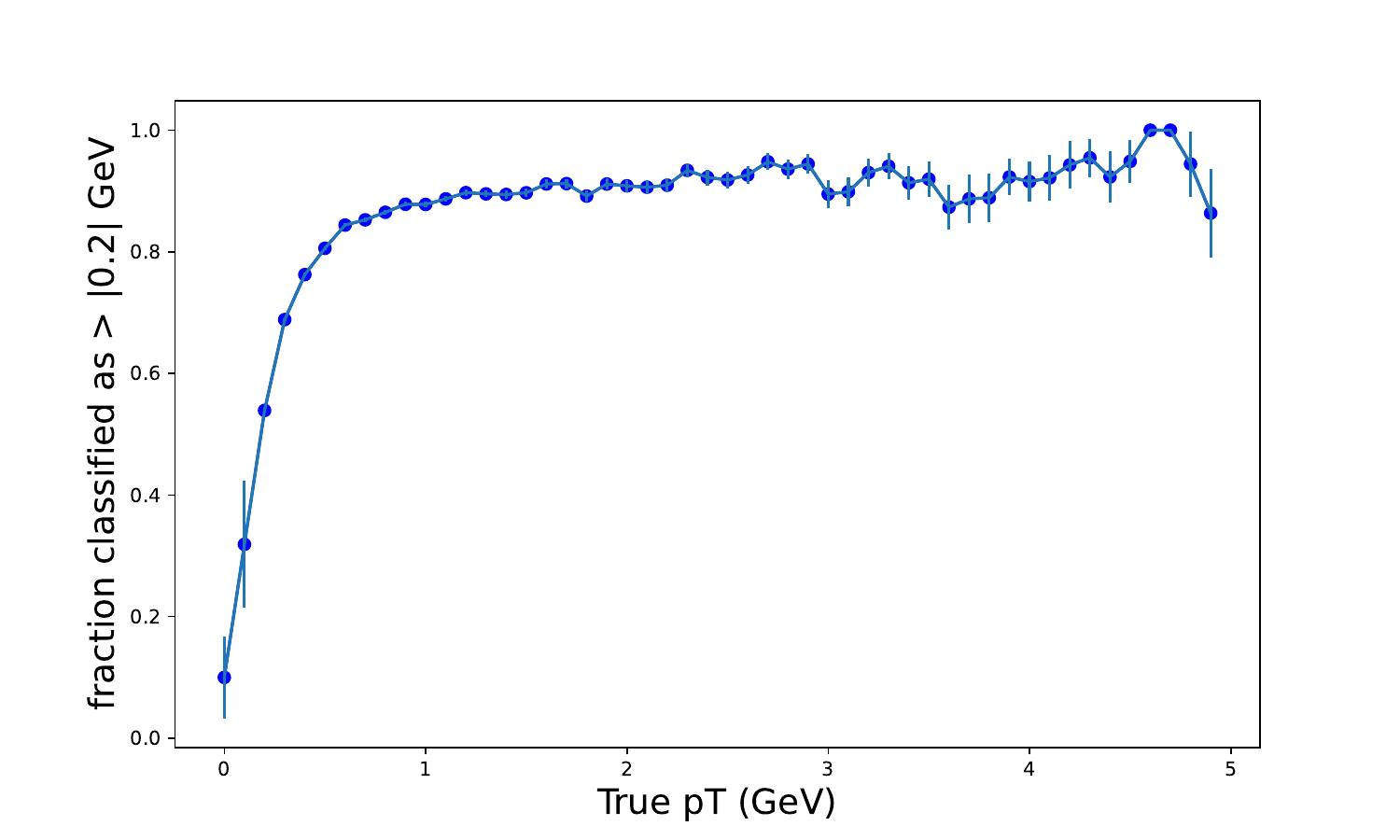}
    \caption{Performance of the best SNN across the full range of $p_T$ values.}
    \label{fig:turn-on-curve}
\end{figure}
From our experiments run so far, we were able to achieve the optimal point between efficiency and reduction with an SNN trained with penalty based fitness function, spatial data reduction scheme of row-stride of 26, $pT$-cutoff of $0.5$, and time-resolution of $200\,$ps. The resulting network had 84 neurons and 423 synapses. The number of tunable parameters in the SNN are 84 threshold values, and 423 of each synaptic weight and delays, giving a total of $930$ parameters. The network had a signal efficiency of $91.89\%$, and data reduction metric of $26.46\%$. Figure~\ref{fig:turn-on-curve} shows the performance of this SNN trained on the set of positively charged clusters.

% \begin{itemize}
%     % \item Table of Deffe hyperparameters
%     \item Deffe results across different spatial encoding schemes  -- Figure
%     \item Turn on curves for the best performing network --Figure
%     \item Compare DNN results and SNN results -- model accuracy, model size

\subsection{DNN results}
The prototype deep neural network classifier to filter single pixel clusters was based on an algorithm using the $y$ coordinate position (1 feature) and the cluster $y$ profile information (13 features). The $y$ coordinate position references the place on the sensor module where a track hits. Cluster $y$ profile refers to the cluster shape summed over pixel rows after 4 nanoseconds. This $y$ profile information is relevant since it is sensitive to the track's incident angle and thus its $p_T$. 

The original neural network model was composed of one dense layer with 128 neurons and 2049 parameters. y-profile model inputs were quantized to 6-bits, and model weights and activations were quantized using the QKeras library.

Two key metrics for evaluating the model are signal acceptance efficiency and data reduction rate. Signal acceptance efficiency is defined as the percentage of tracks with  $p_T$ greater than 2 GeV that were accurately classified. The signal acceptance efficiency of the full precision model was 94.8 $\%$, whereas that of the quantized model was 91.7$\%$. The data reduction rate when implementing the full precision model would be 24.08$\%$, while that of the quantized model would be 25.72$\%$. 

% note: i see above that you used only positive tracks? whereas we used positive and negative; or did you mean you used only "unflipped" tracks?
A more complex model that included y-profile information from the first eight timeslices, as well as both positive and negative tracks was also implemented to explore possible gains from incorporating timing information. Signal efficiency was found to increase to $98.0\%$. Power and resource constraints currently prevent the implementation of such a DNN-based timing model on current chip architecture, thus opening the window for SNN models.

% \end{itemize}

\section{Discussion}
\label{sec:disc}
The smart-pixel application presents us an opportunity to explore the neuromorphic co-design of for a real-world scientific problem of designing an efficient, low-latency and accurate on-sensor hardware. In addition to the model's algorithmic performance, the model size and latency of the underlying hardware are also crucial in the instrumentation designed for scientific experiments. Table~\ref{tab:comparison} compares the performance of our EONS trained SNN with a feed-forward deep neural network described in the previous section. The DNN has a feed-forward and layered architecture, while the SNN trained by EONS has an unstructured any-to-any connectivity \cite{schuman2020evolutionary}. It can also be seen that while the SNN compares well with DNN in terms of signal efficiency, the reduction rate needs improvement. 
We plan to investigate different algorithmic improvements and encoding schemes to improve the two metrics of evaluating the SNN for the smart-pixel system, especially in similar problems with imbalanced data available for training.

However, it can be noted that the SNN takes into account the timing aspect of the incoming sensor charge values. Overall, number of parameters is nearly half compared to the DNN, hence, has great potential to be realized for on-sensor processing. Also, in terms of spatial data reduction, it is to be noted that y-profile used in DNN and the row-stride scheme used in SNNs are similar, since, both consider the projection of data along the y-axis. However, in SNNs the spikes along columns are 'OR'ed rather than summed, as is the case in the DNN, thereby, also reducing the number of operations needed in the pre-processing stage. The input temporal encoding of data into spikes would be built-into the design of the analog front-end of the readout hardware. Hence, there is potential for co-designing a memory and energy efficient hardware in neuromorphic computing. This also opens up opportunities for exploring the prospects of low-precision neuromorphic hardware designs in CMOS and other nanoscale memory based arrays.

\begin{table}[]
\caption{Performance of SNN and DNN models}
\label{tab:comparison}
\begin{tabular}{llll}
\hline
\textbf{Models}                                                       & \begin{tabular}[c]{@{}l@{}}DNN \\ (full\\ precision)\end{tabular} & \begin{tabular}[c]{@{}l@{}}DNN (quantized\\ (5-bit weights+\\ 10-bit activations)\end{tabular} & \begin{tabular}[c]{@{}l@{}}SNN \\ (this work)\end{tabular} \\ \hline
\textbf{\begin{tabular}[c]{@{}l@{}}Signal \\ Efficiency\end{tabular}} & 94.8\%                                                           & 91.7\%                                                                                        & 91.89\%                                                    \\
\textbf{\begin{tabular}[c]{@{}l@{}}Data \\ Reduction\end{tabular}}    & 24.02\%                                                             & 25.72\%                                                                                          &  25.47\%                                                \\
%\textbf{\begin{tabular}[c]{@{}l@{}}Data \\ Reduction\end{tabular}}    & 24.02\% 51.1\%                                                            & 25.71\% 51.8\%                                                                                         &  25.47\% %26.46\%                                                    \\
\textbf{Neurons}                                                      & 128                                                               & 128                                                                                            & 84                                                         \\
\textbf{Parameters}                                                   & 2049                                                              & 2561                                                                                           & 930                                                        \\ \hline
\end{tabular}
\end{table}

\section{Summary}
\label{sec:summary}
We have presented our initial approach towards applying neuromorphic SNN model for processing sensor data for high energy physics experiments, by means of encoding the sensory charge waveform into trains of spikes and training an SNN for a binary classification task on this spike data. Further, we have also shown that with several spatial data reduction schemes with a simple 'OR' operation also aids in model's performance in addition to reducing the time-to-solution due to reduced network size. The SNN classifier is trained using EONS evolutionary algorithm on a limited dataset captured from a specific region of the sensor array and positively charged samples. On this set, our SNN shows signal efficiency of around $91.0\%$ as seen in the turn-on curve plot in Figure~\ref{fig:turn-on-curve}, which is comparable with that of a deep neural network. The best optimized SNN is able to make the classification decision within 20 timesteps with binary-valued spikes as the activation data, hence, it holds greater potential for designing low power and low latency hardware for the instrumentation requirements of the HL-LHC system.

Going forward, we would be investigating the SNN training and data encoding schemes for the entire range of samples in the sensor array. This study also provides for an exemplary case for neuromorphic co-design process for several edge and scientific applications bringing the compute closer to sensors. Neuromorphic computing holds great potential in developing algorithm informed by the underlying hardware and also in the other direction of carrying out research in new devices and architectures that can leverage the computational energy-efficiency benefits provided by event-driven processing within SNNs.

\begin{acks}
% Funding -- ASCR DOE, Abisko Project
This manuscript has been authored in part by UT-Battelle, LLC under Contract No. DE-AC05-00OR22725 with the U.S. Department of Energy. The United States Government retains and the publisher, by accepting the article for publication, acknowledges that the United States Government retains a non-exclusive, paid-up, irrevocable, world-wide license to publish or reproduce the published form of this manuscript, or allow others to do so, for United States Government purposes. The Department of Energy will provide public access to these results of federally sponsored research in accordance with the DOE Public Access Plan (http://energy.gov/downloads/doe-public-access-plan). %This research used resources of the Oak Ridge Leadership Computing Facility, which is a DOE Office of Science User Facility supported under Contract DE-AC05-00OR22725.
This research used resources of the Experimental Computing Laboratory (ExCL) at the Oak Ridge National Laboratory, which is supported by the Office of Science of the U.S. Department of Energy under Contract No. DE-AC05-00OR22725.

This work was funded in part by the DOE Office of Science, Advanced Scientific Computing Research (ASCR) program.
This research is funded by the DOE Office of Science Research Program for Microelectronics Codesign (sponsored by ASCR, BES, HEP, NP, and FES) through the Abisko Project with program managers Robinson Pino (ASCR). Hal Finkel (ASCR), and Andrew Schwartz (BES).  Personnel were also supported through funding from the NSF Elementary Particle Physics program.
\end{acks}

\bibliographystyle{ACM-Reference-Format}
\bibliography{references}

% \appendix


\end{document}